\documentclass[pmlr]{jmlr}

\usepackage{longtable}
\usepackage{cleveref}

\usepackage{booktabs}
\usepackage[load-configurations=version-1]{siunitx} 

\usepackage{float}

\theorembodyfont{\upshape}
\theoremheaderfont{\scshape}
\theorempostheader{:}
\theoremsep{\newline}

\jmlrvolume{1}
\jmlryear{2021}
\jmlrworkshop{Under Review}

\title[The NeurIPS 2021 NetHack Challenge]{Insights from the NeurIPS 2021 NetHack Challenge}

\newcommand{\first}{$^{\texttt{\large *}}$}
\newcommand{\fair}{$^{\texttt{\large +}}$}
\newcommand{\aicrowd}{$^{\texttt{\large =}}$}
\newcommand{\ucl}{$^{\texttt{\large !}}$}
\newcommand{\oxford}{${\texttt{\large "}}$}
\newcommand{\airi}{$^{\texttt{\large ?}}$}
\newcommand{\sber}{$^{\texttt{\large \#}}$}
\newcommand{\kakaobrain}{$^{\texttt{\large \$}}$}
\newcommand{\chaoticdwarf}{$^{\texttt{\large <}}$}
\newcommand{\independent}{$^{\texttt{\large -}}$}

\makeatletter
\newcommand\blfootnote[1]{%
  \bgroup
  \renewcommand*{\footnoteseptext}{}%
  \renewcommand\thefootnote{}%
  \footnotetext[0]{#1}%
  \egroup
}
\makeatother

\newcommand{\affilnote}{\blfootnote{
\first Lead organizers. Author ordering for the other authors is alphabetical.
\fair Facebook AI Research
\aicrowd AIcrowd
\ucl University College London
\oxford University of Oxford
\sber Sberbank AI Lab
\airi AIRI
\kakaobrain KakaoBrain
\chaoticdwarf University of Eastern Finland
\independent Independent
}}

\author{
  \Name{Eric {Hambro}}\first\fair \Email{ehambro@fb.com} \AND
  \Name{Sharada {Mohanty}}\first\aicrowd \Email{mohanty@aicrowd.com} \AND
  \Name{Dmitrii {Babaev}}\sber\airi \Email{babaev@airi.net} \AND
  \Name{Minwoo {Byeon}}\kakaobrain \Email{minu@kakaobrain.com} \AND
  \Name{Dipam {Chakraborty}}\aicrowd \Email{dipam@aicrowd.com} \AND
  \Name{Edward {Grefenstette}}\fair\ucl \Email{egrefen@fb.com} \AND
  \Name{Minqi {Jiang}}\fair\ucl \Email{msj@fb.com} \AND
  \Name{Daejin {Jo}}\kakaobrain \Email{daejin.jo@kakaobrain.com} \AND
  \Name{Anssi {Kanervisto}}\chaoticdwarf \Email{anssi.kanervisto@uef.fi}  \AND
  \Name{Jongmin {Kim}}\kakaobrain \Email{jmkim@kakaobrain.com} \AND
  \Name{Sungwoong {Kim}}\kakaobrain \Email{swkim@kakaobrain.com} \AND
  \Name{Robert {Kirk}}\ucl \Email{robert.kirk.20@ucl.ac.uk} \AND
  \Name{Vitaly {Kurin}}\oxford \Email{vitaly.kurin@cs.ox.ac.uk} \AND
  \Name{Heinrich {K{\"u}ttler}}\fair \Email{hnr@fb.com} \AND
  \Name{Taehwon {Kwon}}\kakaobrain \Email{taehwan.kwon@kakaobrain.com}  \AND
  \Name{Donghoon {Lee}}\kakaobrain  \Email{dhlee@kakaobrain.com} \AND
  \Name{Vegard {Mella}}\fair  \Email{vegardmella@fb.com} \AND
   \Name{Nantas {Nardelli}}\oxford  \Email{nantas.nardelli@gmail.com} \AND
  \Name{Ivan {Nazarov}}\airi \Email{nazarov@airi.net} \AND
  \Name{Nikita {Ovsov}}\sber \Email{ovsov.n.p@sberbank.ru} \AND
  \Name{Jack {Parker-Holder}}\oxford \Email{jackph@robots.ox.ac.uk} \AND
  \Name{Roberta {Raileanu}}\fair \Email{raileanu@fb.com} \AND
  \Name{Karolis {Ramanauskas}}\independent \Email{karolis.ram@gmail.com} \AND
  \Name{Tim {Rockt{\"a}schel}}\fair\ucl \Email{rockt@fb.com} \AND
  \Name{Danielle {Rothermel}}\fair \Email{drotherm@fb.com} \AND
  \Name{Mikayel {Samvelyan}}\fair\ucl \Email{samvelyan@fb.com} \AND
  \Name{Dmitry {Sorokin}}\airi \Email{sorokin@airi.net} \AND
  \Name{Maciej {Sypetkowski}}\independent \Email{maciej.sypetkowski@gmail.com} \AND
  \Name{Micha\l{} Sypetkowski}\nametag{\independent}\Email{m.sypetkowski@gmail.com}~\hfill
}

\editor{Editor's name}

\newcommand{\NLE}{\texttt{NLE}}

\begin{document}

\maketitle

\begin{abstract}

In this report, we summarize the takeaways from the first NeurIPS 2021 NetHack Challenge.
Participants were tasked with developing a program or agent that can win (i.e., `ascend' in) the popular dungeon-crawler game of NetHack by interacting with the NetHack Learning Environment (\NLE),  a scalable, procedurally generated, and challenging \textit{Gym} environment for reinforcement learning (RL).  The challenge showcased community-driven progress in AI with many diverse approaches significantly beating the previously best results on NetHack. Furthermore, it served as a direct comparison between neural (e.g., deep RL) and symbolic AI, as well as hybrid systems, demonstrating that on NetHack symbolic bots currently outperform deep RL by a large margin.
Lastly, no agent got close to winning the game, illustrating NetHack's suitability as a long-term benchmark for AI research.

\textbf{}



\end{abstract}

\section{Introduction}
\label{sec:intro}
\affilnote

Progress in artificial intelligence research requires benchmarks that test the limits of current methods. To this end, the NeurIPS 2021 NetHack Challenge was a competition to drive the open benchmarking of current sequential decision-making methods on the NetHack Learning Environment~\citep[\NLE{},][]{kuettler2020nethack}.
 \NLE{} is a fully-featured \emph{Gym} environment (\cite{DBLP:journals/corr/BrockmanCPSSTZ16}) based on the popular open-source terminal-based
single-player procedurally-generated ``dungeon-crawler'' game, NetHack (\cite{raymond1987guide, NetHackOrg}).

Aside from procedurally generated content, NetHack is an
attractive research platform as it contains hundreds of enemy and
object types, has complex and stochastic environment dynamics, and
has a clearly defined goal (descend the dungeon, retrieve an
amulet, and ascend) which can be achieved in a diverse set of ways.
The game is considered one of the hardest in the world\footnote{\url{https://www.telegraph.co.uk/gaming/what-to-play/the-15-hardest-video-games-ever/nethack/}}, with winning episodes lasting 100,000s of steps, and a permadeath setting that starts agents at the beginning in a whole new world if they die in the dungeon.
NetHack is even difficult to master for human players who often rely on external knowledge, such as the many extensive community-created documents outlining various strategies for the game (\cite{nhwiki,NetHackSpoilers}), to learn about strategies and NetHack's complex dynamics and secrets.

NetHack has a long history of online tournaments, often played on various competition servers. The longest running of these is the \texttt{/dev/null/nethack} tournament\footnote{\url{https://nethackwiki.com/wiki//dev/null/nethack_tournament}}  which ran every year from 1999 to 2016 and has now been superseded by the November NetHack Tournament.\footnote{\url{https://www.hardfought.org/tnnt/}}  While bots are forbidden from entering this tournament, the community still has a large history of creating symbolic bots, such as SWAGGINZZZ \citeyearpar{swagginzzz}, \citet{taebwikibots}, BotHack \citeyearpar{bothack}.  In the past, these bots have often made progress by exploiting bugs or weaknesses in the earlier versions of the game that have since been removed by the game developers. To the best of our knowledge, no bot has ever ascended in the most recent version of the game, NetHack 3.6.6.

At NeurIPS 2021, we challenged participants to either train a machine learning agent or hard-code a symbolic bot to win the game of NetHack.
In this report, we present in-depth analyses of the main results of this challenge, in particular
i) the community-driven progress in AI for NetHack compared to the previous state-of-the-art, ii) the current dominance of symbolic bots over deep RL approaches, and iii) the difficulty and complexity of NetHack leading to none of the participating agents getting close to winning the game.

\section{Competition}
\label{sec:competition}
In this section, we describe the \NLE{} environment to which participants had access to, as well as the evaluation metrics and competition structure.

\subsection{Environment}
\label{sec:Environment}


We created the new \NLE{} task, ``\texttt{challenge}'', for the competition, in order to expose the full game of NetHack 3.6.6 in all its complexity. This environment expanded the action space from 23 to 113 actions (the full keyboard), removed the autoclosing of popups, and allowed all forms of input modality. It expanded the observation space to be as broad as possible, including structured observations like \texttt{glyphs}, \texttt{message} and \texttt{blstats}, inventory observations like \texttt{inv\_strs} and \texttt{inv\_glyphs}, as well as the raw terminal outputs in \texttt{ttychars}, \texttt{ttycolors}, \texttt{ttycursor} which is what human players have access to when playing NetHack. Finally, this environment enforced rotation of the starting character's race, role, alignment and gender.  This in particular presented a much harder challenge to competitors than the previous character-specific NetHack score task in \citep{kuettler2020nethack}, thus incentivising the development of general agents instead of character-specific approaches.

We released \texttt{challenge} in \NLE{} v0.7.0, and improved it with minor bug fixes over the first month of the challenge, before freezing at v0.7.3. All evaluations were run on v0.7.3.


\subsection{Metrics}
\label{sec:metrics}

Submission performance was measured with a tuple of three statistics calculated over the test episodes.
These statistics were, in order of tie-breaking: number of ascensions, median in-game score, and mean in-game score.
While the in-game score does not directly correspond to making progress towards ascension, it does generally correspond to good play. As evidence, the minimum score needed for ascension is 12,200, but the average ascending run score is 6.98 million.\footnote{Calculated from a sample of 20,000 ascending human runs from alt.org} However, this correlation does not always hold---some expert players attempt to ascend while, in fact, minimizing the score. Full details of the score methodology can be found on the NetHack Wiki.\footnote{\url{https://nethackwiki.com/wiki/Score}}



\subsection{Competition Structure}
\label{sec:rules_prizes}

Thanks to our sponsors, Meta AI and DeepMind, the competition was able to award \$20,000 in prizes to competitors, across 4 different tracks. These tracks were: Best Overall Agent, Best Neural Agent,\footnote{Referred to as "Best Agent Making Substantial Use of a Neural Network" on aicrowd.com} Best Symbolic Agent,\footnote{Referred to as "Best Agent \textbf{Not} Making Use of a Neural Network" on aicrowd.com} and Best Agent from an Academic Lab/Independent Researcher(s).  This was designed to incentivise a showdown between symbolic and deep RL methods since Tracks 2 and 3 were mutually exclusive, but otherwise, single teams could win multiple tracks.  Given the suitability of \NLE{} as a “computationally cheap” Grand Challenge, we also sought to incentivise submissions from independent researchers or academic labs.

The competition was divided into two phases: a development phase, running June to October, and a test phase, for the last two weeks of October. Only the top 15 entries for each track qualified for the test phase. In the development phase, we evaluated on 512 episode rollouts, within a 2-hour window, and in the test phase, three submissions were evaluated on 4,096 episode rollouts, over a day. This allowed competitors to evaluate different approaches over the bulk of the competition but allowed us to whittle down competitors to a manageable number for a thorough final evaluation.


\subsection{Evaluation Setup}
\label{sec:evaluation_setup}

The AIcrowd platform was the technological centre of the competition.  AIcrowd hosted a starter kit on their GitLab instance, with a TorchBeast~\citeyearpar{torchbeast2019} training pipeline, two pretrained models, and a submission pipeline. They also hosted the leaderboard, a discussion forum and demo Colab notebooks. Most crucially they hosted an automated evaluation service and provided team members with feedback on their performance. More details on the evaluation setup can be found in \Cref{apd:aicrowd}.


\begin{figure}
    \centering
    \includegraphics[width=.8\linewidth]{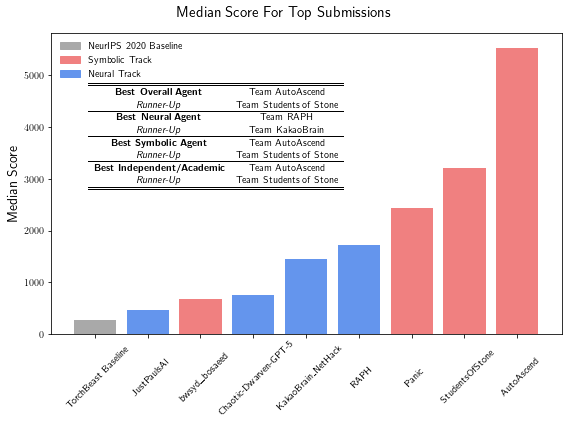}
    \caption{Final Rankings of Top 8 Teams}
    \label{fig:results}
\end{figure}





\section{Results \& Discussion}
\label{sec:results}
In this section, we summarize and discuss the results of the NeurIPS 2021 NetHack Challenge. Overall, the NetHack Challenge ran for 144 days, with 483 entrants registering for the competition, and 631 submissions received over that time.

\subsection{Competition Results}
As mentioned in Section \ref{sec:competition}, the competition was divided into two phases. In Phase 1, 28 entrants beat the official IMPALA baseline, and 17 passed through to Phase 2.  Of these Phase 2 entrants, 8 entrants beat the IMPALA baseline by a significant margin (33\%+), and the rankings and prizewinners are shown in Figure \ref{fig:results}.

The results demonstrate that symbolic approaches currently significantly outperform machine learning, and in particular deep RL-based agents. The top three places in the Overall Best Agent all went to agents from the Symbolic Agent Track, and the winning Neural Agent (RAPH) was a hybrid model, making heavy use of symbolic methods and a neural network. Symbolic bots outperformed other approaches by a significant margin: the best symbolic bot's median score was nearly threefold that of the neural bot's. This lead extends fivefold when looking at the top-scoring episode from each track. Nonetheless, deep reinforcement learning bots made significant progress, offering a nearly fivefold improvement in median score over our NeurIPS 2020 baseline~\citep{kuettler2020nethack}.
Overall, these results demonstrate significant community-driven progress on AI for NetHack.

\subsection{Analyzing Agent Behavior}
To get a better understanding of the results, we analysed the trajectories of all three ``final round" submissions from the Phase 2 finalists.  In particular, we looked at: score distributions by role; dungeon level exploration frequencies; most common causes of death; and other statistics like maximum armor class change, or episode length. More details can be found in Appendix \ref{apd:trajectories}.

We found that the performance of all agents was significantly tied to their starting role and that such performance was often very heavy-tailed (see \Cref{fig:score-apd} in Appendix \ref{apd:trajectories}).
For instance, “easy” starting roles like \textit{Barbarian}, \textit{Monk}, \textit{Samurai} and \textit{Valkyrie} would often have a higher median score than “harder” roles like \textit{Healer}, \textit{Rogue}, \textit{Tourist} and \textit{Wizard}, and also a much higher top score.

Since submissions were heavily incentivised to focus on their below-average characters, we noticed many strategies catered to this directly. We sometimes saw behaviours such as early termination of episodes beyond a certain internal score so as not to run down the clock (see Figure \ref{fig:behaviour-apd} in Appendix \ref{apd:trajectories}), or the intentional restriction of the agent to the top-level dungeons (see Figure \ref{fig:depth-apd} in Appendix \ref{apd:trajectories}).
While the latter behaviour protects weak characters from dangerous monsters, it also likely contributes to the very high incidence of death due to ``fainted of starvation'' (see Figure \ref{fig:death-apd} in Appendix \ref{apd:trajectories}) since agents can run out of food easily if they do not explore the dungeon.


Although the winning agent set a new record for \NLE{}, achieving a median score of ~5,300, this is considered only just above a Beginner\footnote{A ``Beginner" score is defined by the NetHack Wiki to be less than 2,000, or 1,000 for a \textit{Wizard}, according to \url{https://nethackwiki.com/wiki/Beginner}} score. This low result can be explained by the serious challenge posed by playing with difficult roles; outliers in the ``easy" characters performed orders of magnitude better. For instance, 1 in 20 of the winning agent's \textit{Valkyries} would get a score greater than 30,000, descend to dungeon level 10, and get to experience level 10 (typically not all in the same run). However, this is still far short of ascension, which requires the completion of a quest, around 50 dungeon levels, and the accumulation of various special items. This gap demonstrates the difficulty of designing agents that can deal with all the dangers in the complex world of NetHack and win the game.

\subsection{Symbolic vs. Neural Approaches}

The NetHack challenge was designed to incentivise competition between neural and symbolic agents.
This year, symbolic methods won, with notable advantages in multiple areas.


Participants building symbolic bots found it easy to define ‘strategy’-like subroutines, thereby incorporating their domain-knowledge of NetHack into their bots. These bots were often equipped with elaborate routines for deciding when to exercise certain behavior based on rich, human-legible representations of the game state.  Such large subroutines were found in all the top symbolic entries, which often had explicit strategies in code like ``Find Minetown''. This strategic play is very useful in the complex world of NetHack.
In contrast, neural agents struggled in the area since hierarchical RL is still an open research problem. It is hard for agents to discover ‘strategy’-like behaviour patterns in environments with a large action space and sparse reward. Moreover, it is unclear how to best provide RL agents with human prior knowledge about such strategies, for example, information from the NetHack Wiki (like armour strength or monster abilities). Lastly, symbolic bots excelled at leveraging long-term relationships in the game to contextualize and refine their behavior. In contrast, learning long-term relationships and performing credit assignment are still open problems in deep RL.

\section{Approaches}
\label{sec:Approaches}

In the following section, we first present the baselines made available to competitors, and then a description of the best performing symbolic, hybrid and neural entries by the teams themselves.

\subsection{Baselines}

Two baselines were made available to competitors over the course of the competition, specifically for the neural track. These baselines formed the basis of the vast majority of submissions, which were all high-throughput model-free policy gradient methods. No baselines were prepared for the symbolic track, meaning that all such solutions were coded from scratch, often spanning to many thousands of lines of code.

\subsubsection{TorchBeast Baseline}
\label{sec:baseline_torchbeast}

\begin{figure}
    \centering
    \begin{minipage}{0.45\textwidth}
        \centering
        \includegraphics[width=\textwidth]{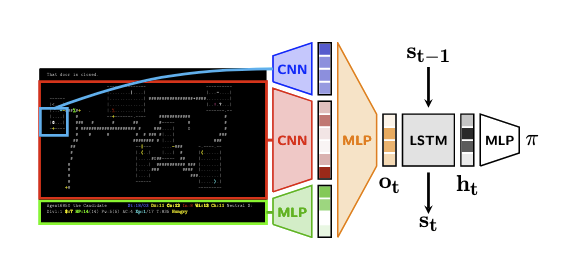} 
        \caption{Neural network model used by the TorchBeast baseline}
        \label{fig:torchbeast}
    \end{minipage}\hfill
    \begin{minipage}{0.5\textwidth}
        \centering
        \includegraphics[width=\textwidth]{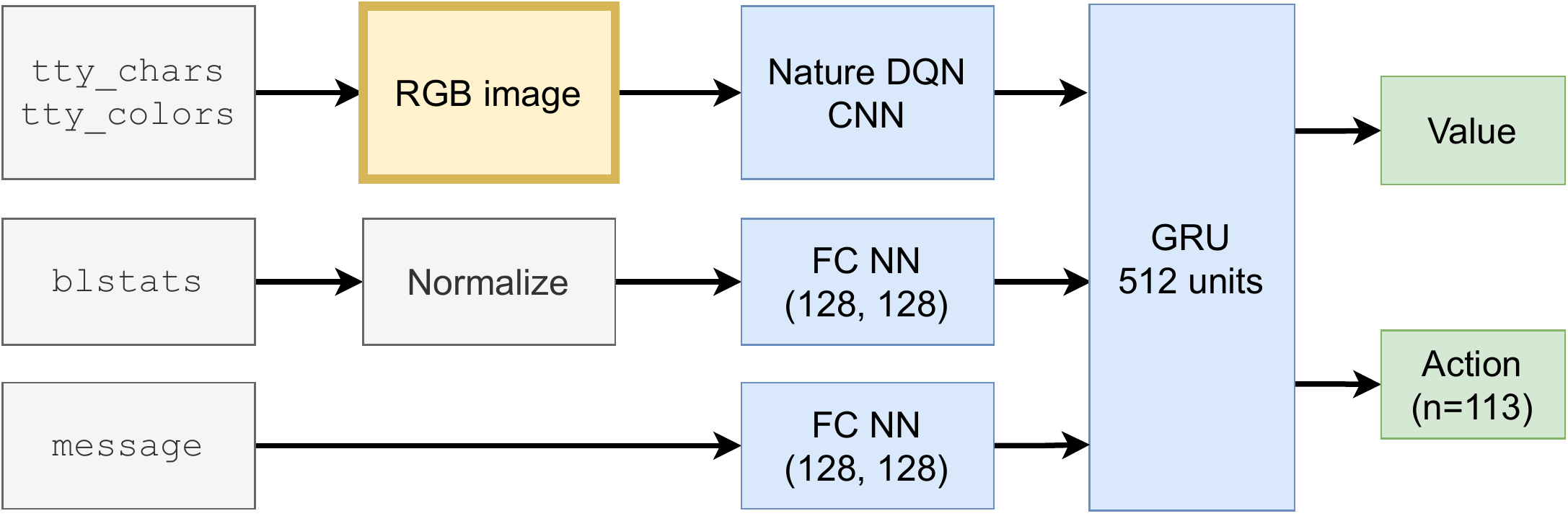} 
        \caption{Neural network model used by the Sample Factory baseline.}
        \label{fig:sf-model}
    \end{minipage}
\end{figure}

The TorchBeast baseline is a modified version of the method presented in the original NeurIPS 2020 paper \citep{kuettler2020nethack}, which is an implementation of IMPALA~\citep{espeholt2018impala} in the open-sourced RL framework TorchBeast \citeyearpar{torchbeast2019}.
This model uses the \texttt{glyphs}, \texttt{blstats}, and \texttt{message} observations as inputs. The \texttt{glyphs} are factorised and embedded before passing into a CNN, while the \texttt{message} is encoded with a character CNN and the \texttt{blstats} is encoded with a feedforward neural network. These are all encoded and passed to an LSTM, that produces baseline and policy heads, albeit in a restricted action space. The architecture is shown in Figure \ref{fig:torchbeast}. This baseline was released with the starter-kit\footnote{\url{https://gitlab.aicrowd.com/nethack/neurips-2021-the-nethack-challenge}} along with versions of the model trained for 0.25B and 0.5B~steps.

\subsubsection{Sample Factory Baseline}


During the competition, the team `Chaotic Dwarven GPT-5' open-sourced\footnote{See \url{https://github.com/Miffyli/nle-sample-factory-baseline}, maintained by co-authors Anssi Kanervisto and Karolis Ramanauskas.} a new baseline solution based on the Sample Factory \citep{petrenko2020sf} framework with asynchronous PPO and a RNN layer. This baseline is more performant than the TorchBeast baseline, reaching a mean score of approximately 700 (vs 400) with under 24 hours of training on a single RTX 2080Ti with 480 parallel \NLE{} environments. This improved performance further lowered the barrier of entry for participants with weaker machines or fewer resources.

Instead of embedding the glyphs, this baseline renders the state as an RGB image,
similar to what a human player sees on a cropped terminal screen.
While this discards important details,
like the specific monster types,
the use of traditional convolutional layers speeds up training and allows agents to learn the associations of colors and shapes. Figure~\ref{fig:sf-model} details the network structure, which also uses \texttt{blstats} and the \texttt{message} observation.

One shortcoming of this baseline is the inability to learn the complex behaviour required by roles such as \textit{Tourist}, \textit{Healer} and \textit{Wizard}, for which it achieves a low score. Another shortcoming is that the learning process is relatively slow as the agent is often stuck exploring different menus instead of making progress on the game itself.

\subsection{Competitor Approaches}
\label{sec:competitors}

We now present the best symbolic, hybrid and neural approaches, described by the teams in their own words.


\subsubsection{Team AutoAscend: Symbolic}
\label{sec:team_autoascend}

\newcommand*{\eg}{e.g.\@\xspace}
\makeatletter
\newcommand*{\etc}{%
    \@ifnextchar{.}%
        {etc}%
        {etc.\@\xspace}%
}




There are two main parts to our framework:

\paragraph{Parsing and maintaining the knowledge about the game state.}
    This consists of parsing ASCII pop-ups, remembering dungeon level state including non-current levels,
    using commands like \texttt{\#terrain} to check covered objects by items and seen traps,
    keeping track of monsters including peaceful monsters, inventory and contents of the hero's bag, \etc.

\paragraph{Decision making.}
    We implemented a custom variation of a behaviour tree capable of dynamic tree expansion, including recursions.
    Strategies are selected based on the current game state.


\NLE{} actions correspond to single keystrokes and therefore their behaviour depends on the context.
For instance, to throw a projectile one usually needs to press 3 keys: `\texttt{t}' to initiate the throw action, a letter representing an item in the inventory, and a letter representing a direction.
We implemented an additional level of abstraction to wrap many of such actions to make the solution more robust and easier to develop.

We introduced the abstraction of a \textit{strategy} which comprises logic that handles a specific behaviour and usually consists of multiple actions, \eg, move to a given $(x,y)$ position, or solve a Sokoban map.
The idea behind strategies is their non-atomic nature and that they may be interrupted under specific conditions, \eg, the `level exploration' strategy should be interrupted when noticing a monster in favour of the `combat' strategy, which in turn should be interrupted by the `emergency healing' strategy.
Throughout the competition, we developed multiple strategies together with the conditions of when they should activate.

One example of a more complex strategy in our solution is combat, which is priority based.
We defined a simplified action space with $4 \times 8 + 2 = 34$ actions:
\textit{movement}, \textit{melee attack}, \textit{ranged attack}, and \textit{ray wand zap} -- all in 8 directions;
\textit{write ``Elbereth"}, and \textit{wait}.
We implemented a heuristic scoring function that assigns a score to every possible action; the one with the highest priority is subsequently executed.
As a simplification, we always use the best available melee weapon, projectile, launcher, \etc taking into account properties like damage, to-hit bonus and the hero's weapon proficiency.
For future work, there is huge potential in improving the action scoring algorithm with more advanced methods, such as an RL model trained on this simplified action space.


Another example of a strategy that resulted in a significant improvement, especially for non-combat oriented roles, is nutrition management.
We use the following three rules which we found efficient:
(1) eat fresh and safe-to-eat corpses from the ground until ``satiated''; (2)~if there are no such corpses, eat food rations gathered during exploration from the inventory in case the hunger state is ``hungry" or worse; (3) as a last resort, pray when the hunger state is ``fainting'', but no more often than 500~turns after the previous prayer.

A more detailed list of strategies, behaviours, and features is presented in Appendix~\ref{apd:autoascend}, along with an image of the custom visualization and debugging tool we developed.





\subsubsection{Team RAPH: Hybrid}
\label{sec:team_raph}


%

We decided to apply a hierarchical approach to the challenge and construct an agent's policy from basic skills dedicated to solving concrete tasks: eating, fighting, dungeon exploration, and inventory management. The approach closely resembles the options framework of \citet{Sutton1999}, in which a higher-level policy orchestrates the execution of eligible lower-level options until termination. This modularity allows us to build a hybrid neural-algorithmic method, where some skills can be trained, and others -- hard-coded.
In our case, we opted for a simplified higher-level policy and implemented it as a rule-based algorithmic decision system, that executes a lower-level skill on a first-fit basis. The priority and triggers of each skill were designed manually and determined based on our expert knowledge of rogue-likes and essential game AI.

One of the most important and complex skills, which accounts for the bulk of in-game score, is battling monsters. Fighting and combat require a complex policy to assess the surrounding topology of the level and choose an appropriate action: approach, outflank, avoid getting surrounded, decide on a melee or ranged attack, heal, wait or flee.
To this end we train a deep neural RL agent based on the TorchBeast baseline, provided in \citet{kuettler2020nethack}, to learn a policy for the lower-level fighting skill. The agent has a discrete action space containing \emph{eight} actions for directional movement or melee attacks, another \emph{eight} actions for directional ranged attacks, and \emph{three} actions for hard-coded composite controls such as waiting, praying and engraving ``Elbereth'', the latter warding off low-level aggressive monsters.
The neural policy uses hand-crafted features related to the map, monster, and hero's vitals, extracted by the lower level algorithmic dungeon level mapping subsystem of the agent.
We implement other skills as hard-coded algorithmic policies based on graph navigation algorithms and expert knowledge.

To train the agent we construct episodes by \emph{pasting contiguous fragments} with transitions in which there is a \emph{hostile monster within the field-of-view} of the agent. In other words, the steps performed by all other skills are ``fast-forwarded'', which makes the environment a partially observed semi-Markov decision process from the point of view of the agent, \citep{Sutton1999}.
The inference of our RAPH agent works as presented in the Appendix, in algorithm~\ref{alg:raph}. If necessary, we first handle pending NetHack's GUI events, such as responding to multi-part message logs. Next, we parse and update the dungeon representation and extract the features for the neural agent. Finally, we sample actions from either the RL agent or the hard-coded skill, whichever one currently holds control, taking into account the distance to the nearest monster.

Analysis of a trained policy shows that move or melee attack actions are used in 66\% of steps, range attack in 21.6\%, wait in 11\%, ``Elbereth'' in 0.3\%, and pray in 0.1\%.

\subsubsection{Team Kakao Brain: Neural}
\label{sec:team_kakao_brain}

%


Our method is based on the V-trace actor-critic framework, IMPALA \citep{espeholt2018impala}, which we used to optimize a policy network through end-to-end RL training from scratch. Here, we briefly describe our approach from the perspectives of 1) observation encoding, 2) separated action spaces, 3) network structure and 4) role-specific training. We provide more details in Appendix \ref{apd:kakaobrain}.

First, we encode \texttt{messages}, \texttt{glyphs}, and an extended \texttt{blstats} as input observations. We also encode the information of \textit{usable items}, \textit{pickable items}, and \textit{spells}, summarized in Figure \ref{fig:kakaobrain_observation_encoding}, so we can explicitly utilize items and spells. Second, we separate the action space into \textit{action-type}, \textit{direction}, \textit{use-item}, \textit{pick-item}, and \textit{use-spell}. In the original action space, a single action can be used for different meanings according to the game status. For example, key `a' (action 24) is used for applying tools as well as for selecting an item or a spell. Our separated and somewhat hierarchical action space clarifies the meaning of each action and also reduces the whole space to the valid action space at each time.

\begin{figure}[t]
\floatconts
  {fig:kakaobrain_observation_encoding}
  {\caption{Observation Encoding of Team KakaoBrain: one-hot encoding for glyph group and object class, and binary encoding for id.}}
  {\includegraphics[scale=0.35]{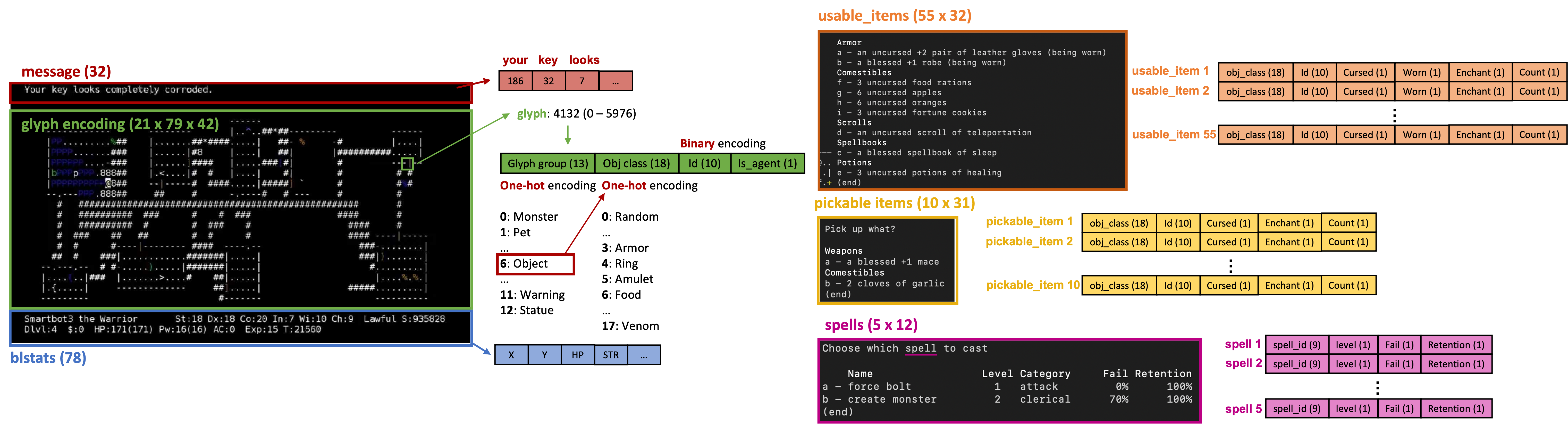}}
\end{figure}

Our network structure is depicted in Figure \ref{fig:kakaobrain_model_structure} in \Cref{apd:kakaobrain}. The encoded \texttt{glyphs} are fed to the CNN while the other observations including tokenized \texttt{messages} and extended \texttt{blstats} are fed to the MLPs. These features as well as the last action type are then concatenated into the GRU network \citep{cho2014properties} for exploiting past information. Moreover, we use Transformer \citep{parisotto2020stabilizing} to combine the GRU output with the information of \textit{usable items}, \textit{pickable items}, and \textit{spells}. The transformer output is then fed to the value head and the multiple policy heads according to the separated action spaces.

To encourage role-specific strategies, we make use of role-specific agents trained by role-specific reward shaping. We maintain such agents for \textit{Healer}, \textit{Ranger}, \textit{Rogue}, \textit{Tourist}, and \textit{Wizard} while the universal agent is used for the other roles. Each policy trains for 10B steps on 4 nodes, with 224 CPU cores and 16 V100 GPUs in total. We modify Sample Factory \citep{petrenko2020sf} for a throughput of 100K frames per second.


\section{Conclusion}
\label{sec:conclusion}

The Challenge produced significant progress in NetHack, including a $5\times$ improvement in deep RL approaches, and a new open-source baseline.
It also provided a valuable point of comparison for neural (deep RL) and symbolic methods, with symbolic bots achieving $3\times$ the median score of deep RL bots in the dungeon.
Notably, the game remains unsolved. The best agents' median score is several orders of magnitude short of a typical (human) `ascension'. As argued in \cite{kuettler2020nethack}, the \NLE{} environment is at the frontier for RL research and it remains to be seen which methods will be able to scale to the point of reliably beating the game.

Future challenges should focus on refining evaluation metrics, such that the ranking of agents better reflects progress in the game. This year many teams indulged in strategies that optimised in-game score at the expense of progressing into the dungeon for ascension. It is possible a new metric, perhaps involving the checkpointing of achievements (as is done for Junethack\footnote{\url{https://nethackwiki.com/wiki/Junethack/FAQ\#Achievements_.2F_Trophies}}) may be more successful in steering agents to ascension.
Future challenges may also seek to proactively encourage teams to exploit external knowledge (\eg, from the NetHack Wiki, or possibly `source-diving' into NetHack's C code), learn from recordings of human play (\eg, \texttt{ttyrec} game replays from alt.org), or test skill acquisition (\eg, using MiniHack \citeyearpar{DBLP:journals/corr/abs-2109-13202}). These are all areas of research that fit naturally into NetHack's ecosystem and pose challenges that would benefit from open competitive benchmarking.

%
%


\acks{The organisers would like to thank Meta AI, DeepMind and AIcrowd for their generous sponsorship of this project.}

\bibliography{jmlrwcp-sample}

\appendix

\section{AI Crowd Evaluation}\label{apd:aicrowd}
The evaluation service on AI Crowd worked as follows: each participant was required to submit their inference code and trained models, using AIcrowd's Gitlab repositories corresponding to their username. The participant’s submitted code was packaged into a Docker image for every submission, and run on AWS “g4dn.xlarge” instances with 4 CPUs, 8 GB of RAM, and 1 NVidia T4 GPU with 16 GB video memory. Simultaneously, AIcrowd also started separate Docker images, which ran the NetHack environment with the fixed settings decided by the organizers, as well as code for tracking the scores of the rollouts. These ran on AWS “t3a.medium” instances with 2 CPUs and 4 GB of RAM. These two Docker images communicated observations and actions to each other respectively and provides the security guarantee that the environment or the scores cannot be hacked in any way by the participants.

\section{Trajectory Analysis}\label{apd:trajectories}

The following plots show the investigation made in the trajectories of the Top 8 finalists in the NetHack Challenge.
In particular, it consists of the three sets of evaluations submitted in Phase 2, consisting of a maximum of 12,288  trajectories. In some cases, this may have been reduced if the agent was unable to complete all 4,096 episodes during an evaluation.

\begin{figure}
    \centering

     \includegraphics[width=.9\linewidth]{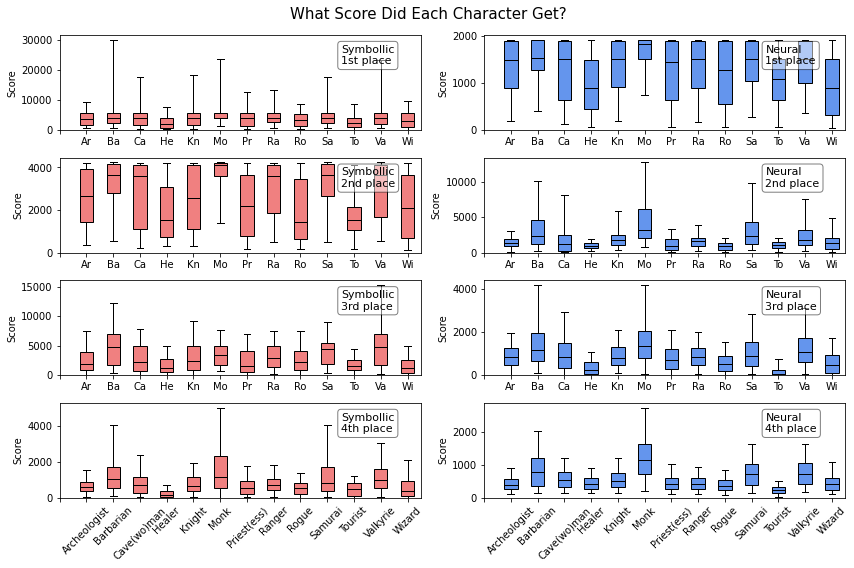}
    \includegraphics[width=.9\linewidth]{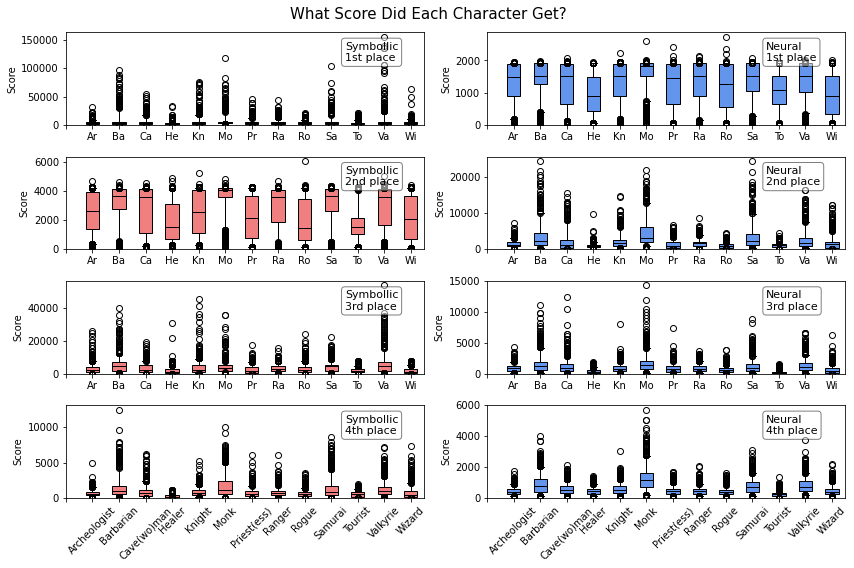}
    \caption{Box plots of the score broken down by starting role. The black line indicates the median, the box-plot is the interquartile range, and whiskers are the 5-95th percentiles. Outliers are shown in the second plot. Note how Symbolic 2nd and Neural 1st show evidence of early episode termination.}
    \label{fig:score-apd}
\end{figure}

\begin{figure}
    \centering
     \includegraphics[width=.9\linewidth]{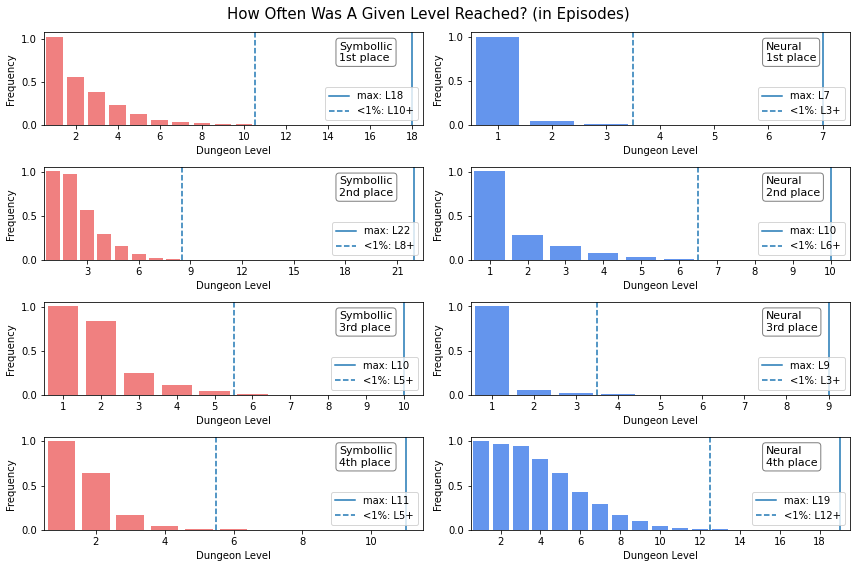}
    \includegraphics[width=.9\linewidth]{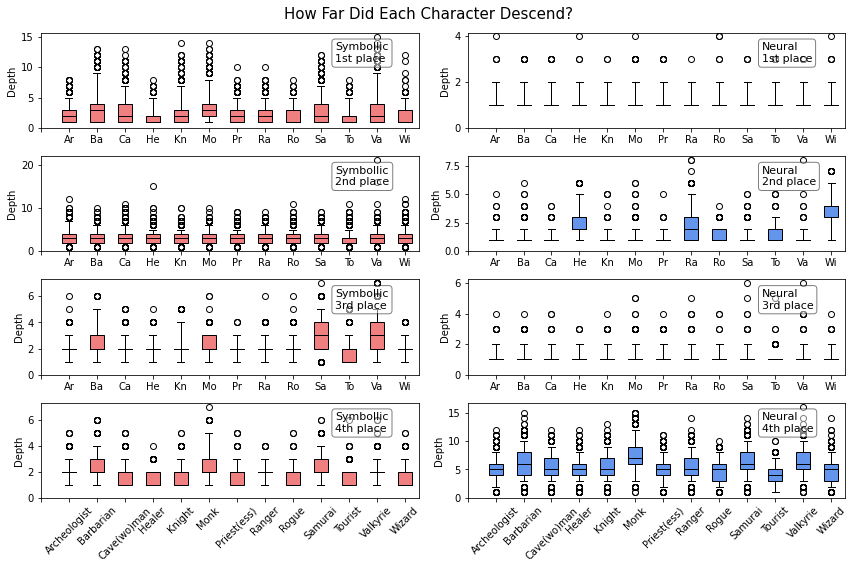}
    \caption{[Top] A plot of dungeon level exploration frequencies per episode. In what fraction of episodes was that dungeon reached?  [Bottom] Top plot broken down by role, plotted as a box plot. The black line indicates the median, the box-plot is the interquartile range, and whiskers are the 5-95th percentiles. Outliers are shown in the second plot.  Note how several entries (Neural 1, Neural 3) indicate restriction to the first level. Characters can still teleport or fall through holes to new levels by accident.}
    \label{fig:depth-apd}
\end{figure}

\begin{figure}
    \centering

     \includegraphics[width=.9\linewidth]{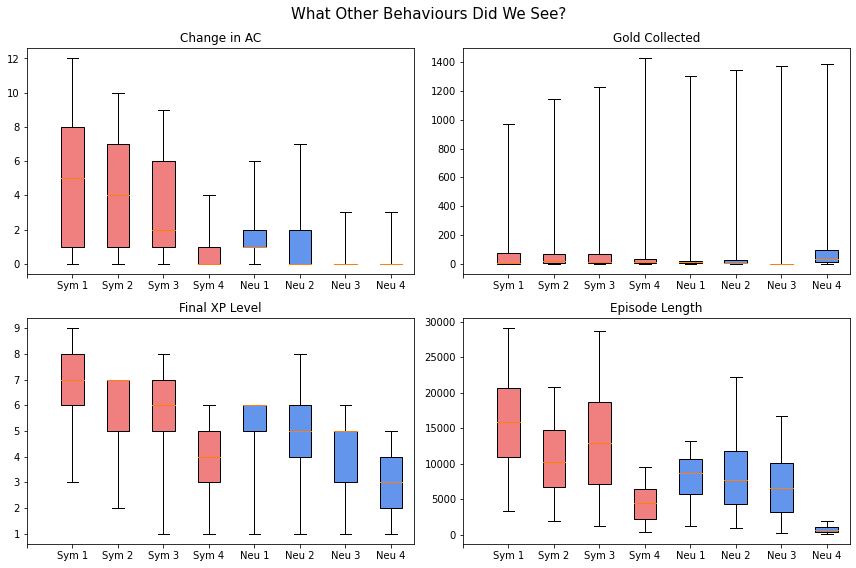}
     \includegraphics[width=\linewidth]{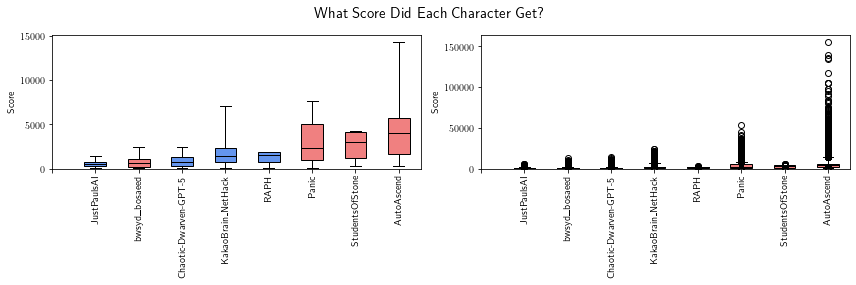}
    \caption{[Top] Boxplots of (clockwise from top left): max change in AC over an episode (NB this can happen by accident by damaging armour or polymorphing); gold accumulated; episode length; experience level at the time of death. [Bottom] The box plots for agent score, aggregated. Left without outliers, right with outliers. The black line indicates the median, the box-plot is the interquartile range, and whiskers are the 5-95th percentiles. Outliers are shown in the second plot. Note how Symbolic 2nd and Neural 1st show evidence of early episode termination.}
    \label{fig:behaviour-apd}
\end{figure}

\begin{figure}
    \centering
     \includegraphics[width=.6\linewidth]{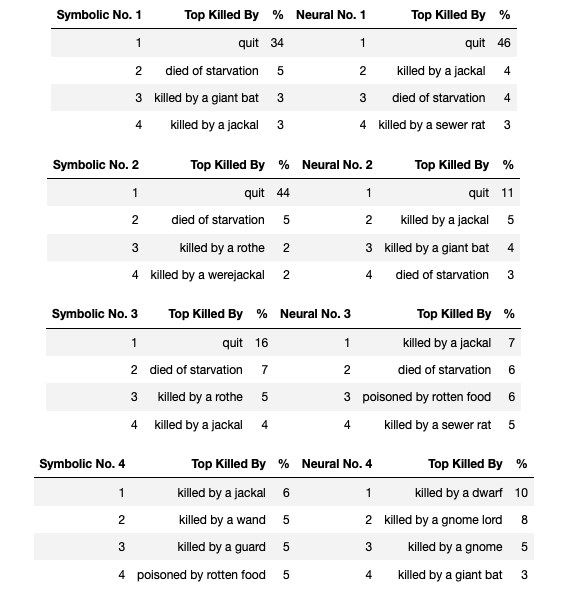}
    \includegraphics[width=.6\linewidth]{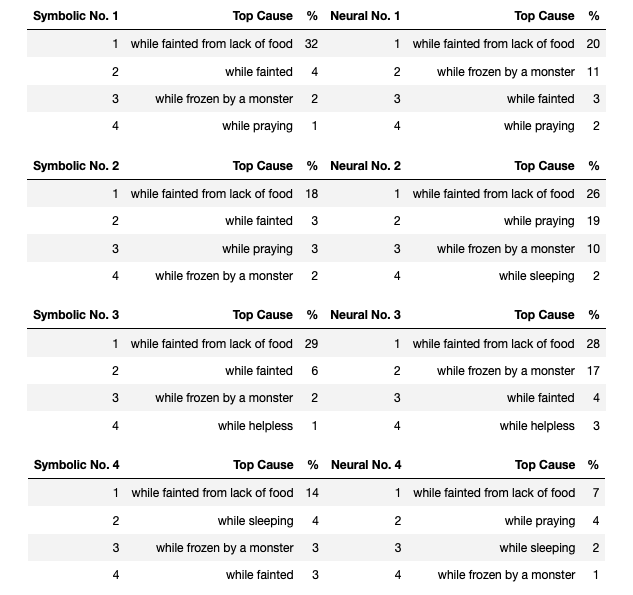}
    \caption{Agents' deaths are described as \textit{``Killed by [Death] while [Cause]''}. In this case [Top] A chart of most common \textit{[Death]}. [Bottom] A chart of most common \textit{[Cause]}.  Note that in many cases, death is aggravated by fainting from a lack of food, or occuring whilst praying.}
    \label{fig:death-apd}
\end{figure}

\begin{figure}
    \centering
     \includegraphics[width=.7\linewidth]{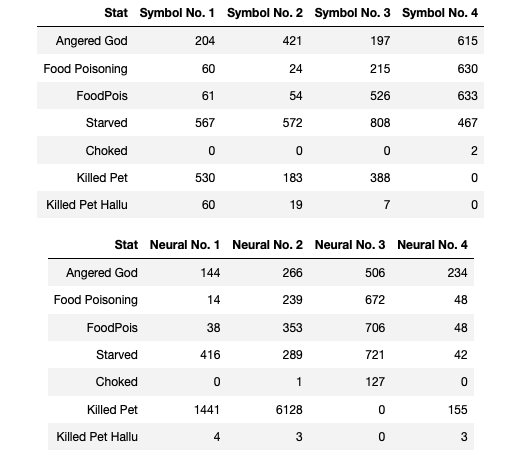}
    \includegraphics[width=.7\linewidth]{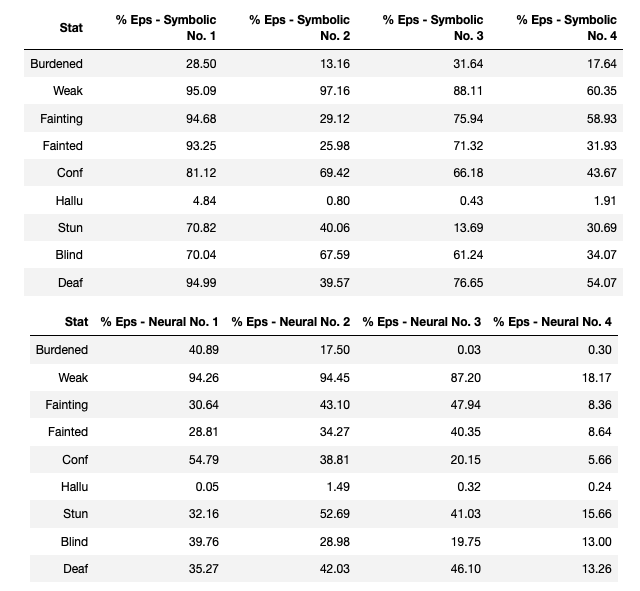}
    \caption{[Top] Number of episodes where certain events took place: death from angered god; death from food poisoning; contracting food poisoned status; starving to death; choking to death; killing a pet; killing a pet while hallucinating. [Bottom] Frequency of episodes where status was encountered.}
    \label{fig:stats-apd}

\end{figure}
\newpage
\section{AutoAscend}\label{apd:autoascend}

\def\apdautoascentitemsep{-0.12em}

We implemented numerous behaviours and features targeting different aspects of the game.
We list the most important ones to show the comprehensiveness of our solution.

\paragraph{Exploration related}
\begin{itemize}
\def\itemsep{\apdautoascentitemsep}
    \item Untrapping traps and chests, looting containers
    \item Keeping track of all dungeon level states including glyphs, items (also stacked), search count per every tile, altars with their alignment, corpse ages, shop positions and types, stairs with the information where they lead
    \item Detecting vault entrances to avoid stepping into them, but if the agent happens to fall into the vault by an accident, drops gold and follows the guard to the exit
    \item Handling map specific behaviors, such as changed diagonal movement in Sokoban, no doors kicking in Minetown, or dungeon level finding helpers, \eg Sokoban is always one level below the Oracle
    \item Basic wish handling
    \item Graceful handling of blindness, confusion, stun, hallucination, polymorph, \etc
    \item Curing lycanthropy by using a sprig of wolfsbane, holy water, or praying
    \item Sokoban solving including error-proof behaviours like checking for monsters before pushing the boulder to a place that cuts the player component in the movement graph, checking for boulder mimics, and destroying boulders
\end{itemize}

\paragraph{Item management related}
\begin{itemize}
\def\itemsep{\apdautoascentitemsep}
    \item Smart item identification, \eg using buy shop prices, possible glyph-object association (\eg gem color), engrave-identification for wands, bijective properties of glyph-object association (if an item is identified, it cannot be under an unidentified glyph), combining results from these methods for drawing better conclusions
    \item Naming items (\texttt{\#call}) for easy instance identification, \eg too old corpses for sacrification, bag identifiers to keep track of items in bags
    \item Using bags for carrying items
    \item Dipping long swords in fountains to get the Excalibur
    \item Identifying BUC status of items using altars
    \item Sacrificing corpses on altars to get an artifact
\end{itemize}

\paragraph{Combat strategy related}
\begin{itemize}
\def\itemsep{\apdautoascentitemsep}
    \item Damage calculation is implemented using NetHack code and the data from the wiki
    \item Enhancing skill proficiencies
    \item Avoiding melee attacking some enemies like ``floating eye" or ``gas spore"
    \item Not using ranged weapons when a pet or a peaceful monster is in the way
    \item Simulating ray wands reflection trajectory (with probabilities for all possible paths)
    \item Healing by using healing potions, using basic healing spells, and praying
\end{itemize}

\begin{figure}
    \centering
    \includegraphics[width=1\linewidth]{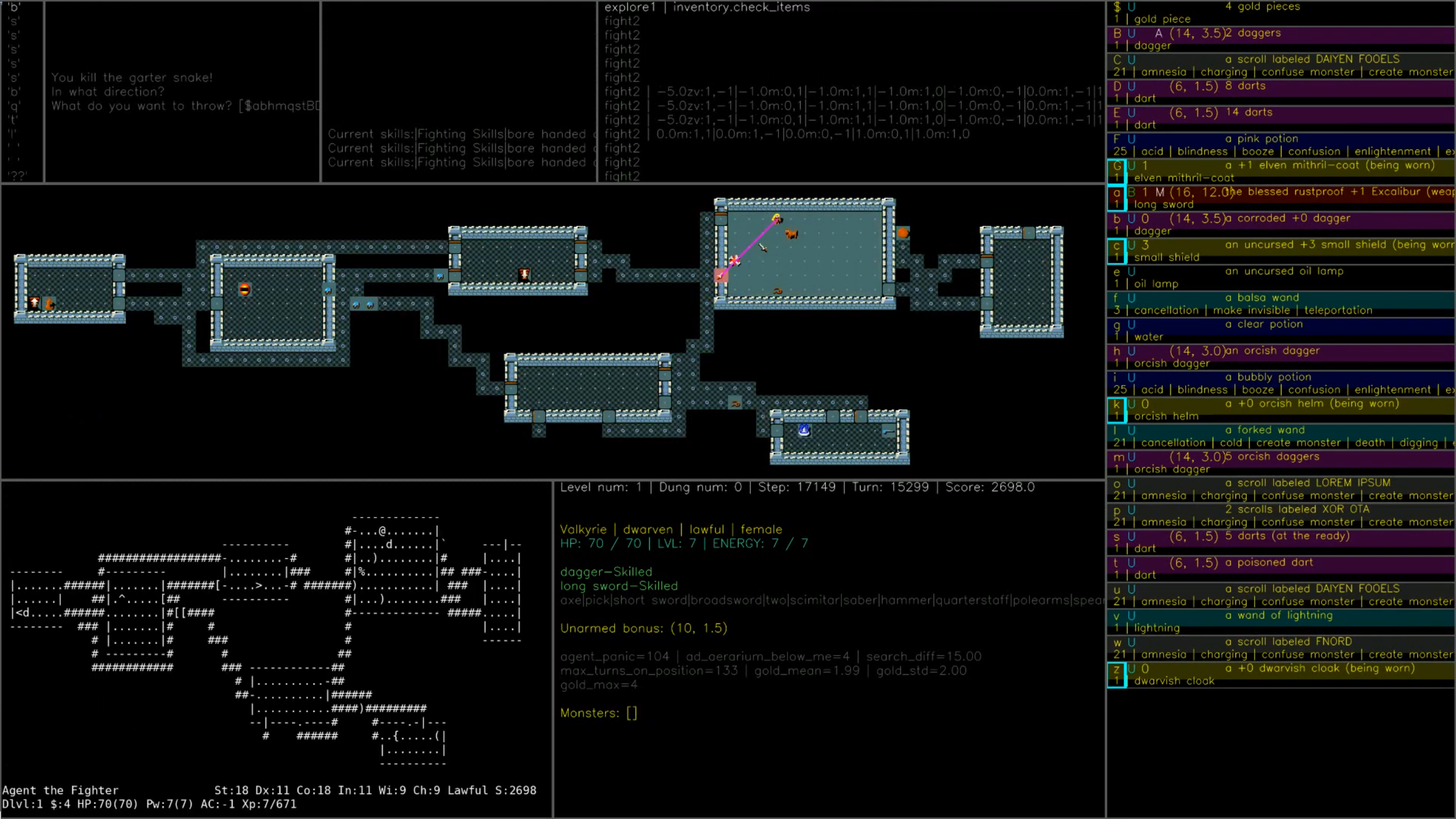}
    \caption{An example visualization of game state and current agent behaviour.
    The state is visualized before every action (in raw \NLE{} action space).
    On the top, there are four log panes (one line per step): raw \NLE{} actions, messages, pop-ups (\eg opened inventory), simplified strategy stacks (from left to right).
    In the centre, the current level is visualized using a graphical tileset.
    On the bottom, there is a raw TTY data visualization and miscellaneous stats.
    The left bar is used for inventory and item knowledge visualization.}
    \label{fig:autoascend_state_vis}
\end{figure}

We also developed a visualisation tool to help us debug the policy of our agent, displayed in Figure \ref{fig:autoascend_state_vis}

\section{RAPH}\label{apd:raph}

\begin{algorithm2e}[H]
\SetKwComment{Comment}{/* }{ */}
\SetKw{Continue}{continue}
\caption{RAPH agent}\label{alg:raph}
\KwData{view\_distance, agent, hard\_coded\_skills}
$state, done \gets env.reset(), False$\;

\While{not done}{
  action\_queue = parse\_message(state)\;

  \If{action\_queue} {
   state, reward, done, info = env.step(action\_queue)\Comment*[r]{We have a prompt to response}
   \Continue
  }

  monster\_distance, preprocessed\_state = parse\_dungeon(state)\;
  \eIf{monster\_distance \textless view\_distance}{
    action\_queue = agent.act(preprocessed\_state)\;
  }{
    action\_queue = first\_fit(hard\_coded\_skills, preprocessed\_state)\Comment*[r]{Select non-rl action on first-fit basis}
  }
  state, reward, done, info = env.step(action\_queue)\;
}
\end{algorithm2e}

\section{KakaoBrain}\label{apd:kakaobrain}
Here we explain details of our approach.
An in-depth model structure is illustrated in \Cref{fig:kakaobrain_model_structure}.
\begin{figure}[t]
\floatconts
  {fig:kakaobrain_model_structure}
  {\caption{Model Structure of Team KakaoBrain.}}
  {\includegraphics[scale=0.35]{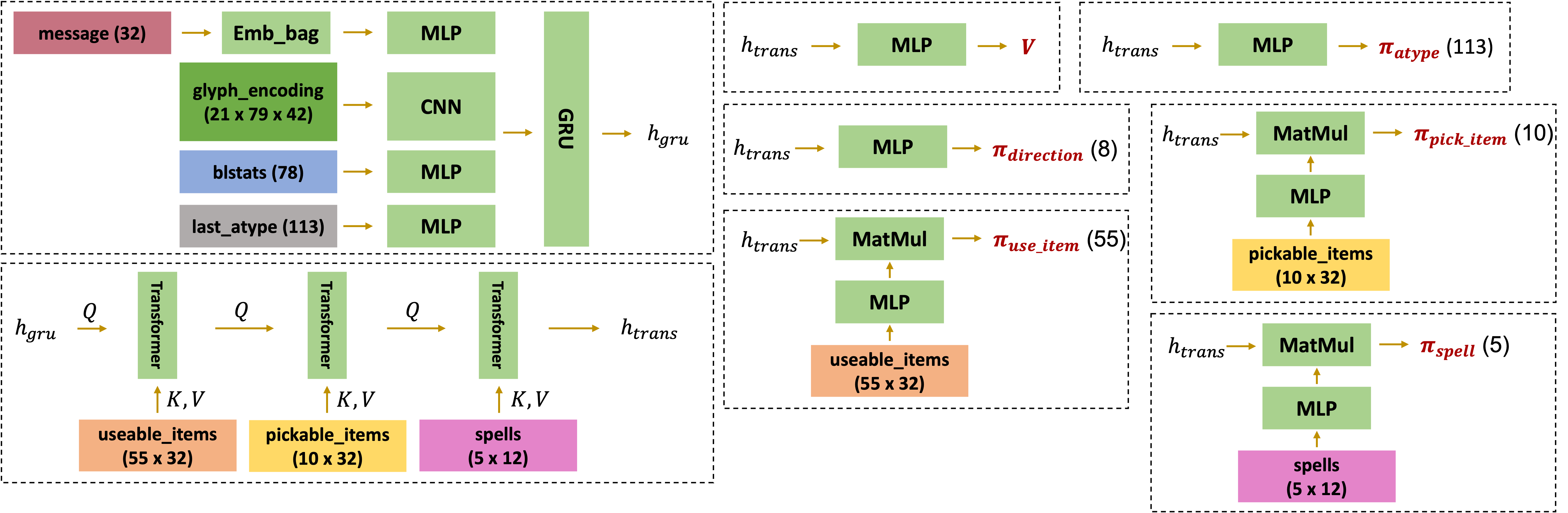}}
\end{figure}
\begin{itemize}
\item \textbf{Observation Encoding}:
For \textit{in-game message tokenization}, we tokenize in-game messages by words rather than characters since it will be easier to understand a sentence in a limited vocabulary setting.
We encode the first 32 words at most (tokens) for a sentence.
For \textit{extended blstats}, we use the blstats with extra information (race, gender, alignment, and condition mask) which helps select optimal behaviors.
For \textit{glyph encoding}, we encode a glyph value (an integer between 0 and 5,976) of each pixel in the screen with its glyph group, object class, id, and is-agent.
We expect that directly providing game-specific information rather than converting to an RGB image is more helpful for the agent to learn a general behavior.
For \textit{usable items}, we encode each item in the inventory with its glyph group, object class, cursed, worn, enchant, and count.
For \textit{pickable items}, we encoded pickable items similarly to usable items by parsing their information from the screen (it is not directly given in the original observation).
For \textit{spells}, we encode each spell information with its id, level, failure, and retention by parsing their information from the screen  (it is not directly given in the original observation). We encode the information of at most 5 spells for simplicity.

\item \textbf{Separated Action Spaces}:
As we explained in the main text, our separated action spaces are composed of \textit{action-type}, \textit{direction}, \textit{use-item}, \textit{pick-item}, and \textit{use-spell}.
\textit{action-type} is the same as the original action space,
\textit{direction} is composed of 8 direction choices for choosing a direction,
\textit{use-item} consists of item choices for choosing an item to be used,
\textit{pick-item} and \textit{use-spell} are comprised of item choices for picking up items and spell choices for casting a spell, respectively.
Generally, the agent chooses an action in action-type, and the other actions are used when required. This encourages better item farming, item utilization, and spell utilization by separating and clarifying action spaces for them.

\item \textbf{Network Structure}:
\textit{Tokenized message} is fed to EmbeddingBag which embeds token id into a vector of 80 size and averages these vectors of multiple (32) tokens.
We use 3-layer CNNs with 128, 64, and 32 channels, stride size of 3, and $2 \times 2$ average pooling for the first and second layers.
We use a 2-layer MLP with the out sizes of 256 for \texttt{blstats}.
We use a 1-layer GRU with the hidden size of 1024.
We use a 1-layer MLP with the out size of 64 for \textit{last atype}.
We use TrXL-I structure without memory \citep{parisotto2020stabilizing} to encode information of usable items, pickable items, and spells.
We use 1-layer TrXL-I with the hidden size of 1024, the head size of 256, and 4 heads for each of them.

\item \textbf{Role-specific Training}:
To improve the score by encouraging a role-specific strategy, we train models that are dedicated to a specific role. In specific, we apply the role-specific reward shaping for Healer, Ranger, Rogue, Tourist, and Wizard.
We set the HP difference as a reward to encourage healing itself for Healer.
We give an incentive in killing monsters with firing for Ranger, killing monsters by throwing for Rogue, killing monsters by throwing for Tourist, and
killing monsters with spells for Wizard.
\end{itemize}

\end{document}